\pdfoutput=1
\documentclass[11pt,a4paper]{article}
\usepackage[table]{xcolor}
\usepackage[hyperref]{ranlp2021}
\usepackage{times}
\usepackage{latexsym}
\usepackage{xcolor}
\usepackage{tikz-dependency}
\usepackage{makecell}

\usepackage{microtype}

\aclfinalcopy % Uncomment this line for the final submission
\title{Data-efficiency of Encodings for Sequence Labeling Dependency Parsing}
\title{Not all Encodings Are Equally Data-hungry in Sequence Labeling Dependency Parsing}
\title{Sequence Labeling Dependency Parsing on Low-resource Languages}
\title{Not All Linearizations Are Equally Data-Hungry\\ in Sequence Labeling Parsing}
\author {\textbf{Alberto Muñoz-Ortiz$^1$}, \textbf{Michalina Strzyz\,$^{1,2}$}, \textbf{David Vilares$^1$} \\
$^1$Universidade da Coruña, CITIC, Spain\\
$^2$Priberam Labs, Portugal\\
\texttt{alberto.munoz.ortiz@udc.es}\\\texttt{michalina.strzyz@priberam.pt}, \texttt{david.vilares@udc.es}\\
}

\date{}

\begin{document}
\maketitle
\begin{abstract}
Different linearizations have been proposed to cast dependency parsing as sequence labeling and solve the task as: (i) a head selection problem, (ii) finding a representation of the token arcs as bracket strings, or (iii) associating partial transition sequences of a transition-based parser to words. Yet, there is little understanding about how these linearizations behave in low-resource setups. Here, we first study their data efficiency, simulating data-restricted setups from a diverse set of rich-resource treebanks. Second, we test whether such differences manifest in truly low-resource setups. The results show that head selection encodings are more data-efficient and perform better in an ideal (gold) framework, but that such advantage greatly vanishes in favour of bracketing formats when the running setup resembles a real-world low-resource configuration.
\end{abstract}

\section{Introduction}

Dependency parsing \cite{mel1988dependency,kubler2009dependency} has achieved clear improvements in recent years, to the point that graph-based \cite{martins2013turning,dozat-etal-2017-stanfords} and transition-based \cite{ma-etal-2018-stack,fernandez-gonzalez-gomez-rodriguez-2019-left} parsers are already very accurate on certain setups, such as English news. In this line, \citet{berzak-etal-2016-anchoring} have pointed out that the performance on these setups is already on par with that expected from experienced human annotators.

Thus, the efforts have started to focus on related problems such as parsing different domains or multi-lingual scenarios  \cite{sato-etal-2017-adversarial,song2019leveraging,ammar2016many}, creating faster models \cite{volokh2013performance,chen-manning-2014-fast}, designing low-resource and cross-lingual parsing techniques \cite{tiedemann2014treebank,zhang-etal-2019-cross}, or infusing syntactic knowledge into models \cite{strubell-etal-2018-linguistically,rotman-reichart-2019-deep}.

This work will lie in the intersection between fast parsing and low-resource languages. Recent work has proposed encodings to cast parsing as sequence labeling \cite{spoustova2010dependency,strzyz-etal-2019-viable,gomez-rodriguez-etal-2020-unifying,li-etal-2018-seq2seq,kiperwasser-ballesteros-2018-scheduled}. This approach computes a linearized tree of a sentence of length $n$ in $n$ tagging actions, providing a good speed/accuracy trade-off.
Also, it offers a naïve way to infuse syntactic information as an embedding or feature \cite{ma-etal-2019-improving,wang-etal-2019-best}. Such encodings have been evaluated on English and multi-lingual setups, but there is no study about their behaviour on low-resource setups, and what strengths and weaknesses they might exhibit.

\paragraph{Contribution} We study the behaviour of linearizations for dependency parsing as sequence labeling in low-resource setups. First, we explore their data efficiency, i.e. if they can exploit their full potential with less supervised data. To do so, we simulate different data-restricted setups from a diverse set of  rich-resource treebanks. Second, we shed light about their performance on truly low-resource treebanks. The goal is to determine whether tendencies from the experiments in the previous phase hold when the language is truly low-resource and when secondary effects of real-world low-resource setups, such as using predicted part-of-speech (PoS) tags or no PoS tags, impact more certain types of linearizations.

\section{Related work}

Low-resource parsing has been explored from perspectives such as unsupervised parsing, data augmentation, cross-lingual learning, or data-efficiency of models. For instance, on unsupervised parsing, \citet{klein-manning-2004-corpus} and \citet{spitkovsky-etal-2010-baby} have worked on generative models to determine whether to continue or stop attaching dependents to a token, while others \cite{le-zuidema-2015-unsupervised,mohananey-etal-2020-self} have studied how to use self-training for unsupervised parsing.

On data augmentation, \citet{mcclosky-etal-2006-reranking} used self-training to annotate extra data, while others have focused on linguistically motivated approaches to augment treebanks. This is the case of \citet{vania-etal-2019-systematic} or \citet{dehouck-gomez-rodriguez-2020-data}, who have proposed methods to replace subtrees within a given sentence. 

On cross-lingual learning, authors such as \citet{sogaard-2011-data} or \citet{mcdonald-etal-2011-multi} trained delexicalized parsers in a source  rich-resource treebank, which are then used to parse a low-resource target language. \citet{falenska-cetinoglu-2017-lexicalized} explored lexicalized versus delexicalized parsers and compared them on low-resource treebanks, depending on factors such as the treebank size and the PoS tags performance. \citet{wang-eisner-2018-synthetic} created synthetic treebanks that resemble the target language by permuting constituents of distant treebanks. \citet{naseem-etal-2012-selective} and \citet{tackstrom-etal-2013-target} tackled this same issue, but from the model side, training on rich-resource languages in such way the model learns to detect the aspects of the source languages that are relevant for the target language. Recently, \citet{mulcaire-etal-2019-low} used a LSTM to build a polyglot language model, which is then used to train on top of it a parser that shows cross-lingual abilities in zero-shot setups.

On data-efficiency, research work has explored the impact of the use of different amounts of data, motivated by the lack of annotated data or by the lack of quality of it. For instance, \citet{lacroix-etal-2016-frustratingly} showed how a transition-based parser with a dynamic oracle can be used without any modifications to parse partially annotated data. They found that this setup is useful to train low-resource parsers on sentence-aligned texts, from a  rich-resource treebank to an automatically translated low-resource language, where only precisely aligned tokens are used for the projection in the target dataset. \citet{lacroix-etal-2016-cross} studied the effect that pre-processing and post-processing has in annotation projection, and concluded that quality should prevail over quantity. Related to training with restricted data, \citet{anderson-gomez-rodriguez-2020-distilling} showed that when distilling a graph-based parser for faster inference time, models with smaller treebanks suffered less. \newcite{dehouck-etal-2020-efficient} also distilled models for Enhanced Universal Dependencies (EUD) parsing with different amounts of data, observing that less training data usually translated into slightly 
lower performance, while offering better energy consumption. \citet{garcia2018new} showed, in the context of Romance languages, that peeking samples from related languages and adapting them to the target language is useful to train a model that performs on par with one trained on fully (but still limited) manually annotated data. Restricted to constituent parsing, \citet{shi-etal-2020-role} analyzed the role of the dev data in unsupervised parsing. They pointed out that many unsupervised parsers use the score on the dev set as a signal for hyper-parameter updates, and show that by using a handful of samples from that development set to train a counterpart supervised model, the results outperformed those of the unsupervised setup. Finally, there is work describing the impact that the size of the parsing training data has on downstream tasks that use syntactic information as part of the input \cite{sagae-etal-2008-evaluating,gomez2019important}.

\section{Preliminaries}

In what follows, we review the existing families of encodings for parsing as sequence labeling (\S \ref{section-encodings}) and the models 
that we will be using (\S \ref{section-framework}).

\subsection{Encodings for sequence labeling dependency parsing}\label{section-encodings} 

\begin{figure}
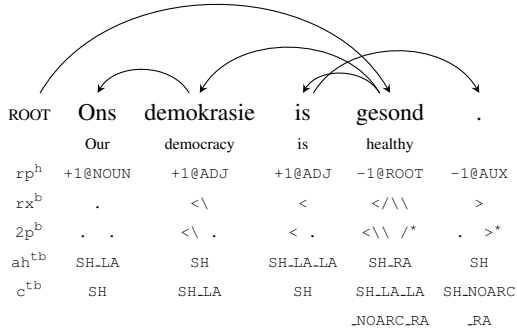

\begin{dependency}[hide label, arc edge]
    \begin{deptext}
        \tiny{ROOT}\&\small{Ons}\&\small{demokrasie}\&\small{is}\&\small{gesond}\&\small{.}\\
        \&\tiny{Our}\&\tiny{democracy}\&\tiny{is}\&\tiny{healthy}\&\\
        \tiny{\texttt{rp\textsuperscript{h}}}  \& \tiny{\texttt{+1@NOUN}} \&\tiny{\texttt{+1@ADJ}}  \& \tiny{\texttt{+1@ADJ}} \& \tiny{\texttt{-1@ROOT}} \& \tiny{\texttt{-1@AUX}} \\
        \tiny{\texttt{rx\textsuperscript{b}}} \& \tiny{\texttt{.}} \& \tiny{\texttt{<\symbol{92}}}  \& \tiny{\texttt{<}} \& \tiny{\texttt{</\symbol{92}\symbol{92}}} \& \tiny{\texttt{>}} \\
        \tiny{\texttt{2p\textsuperscript{b}}} \& \tiny{\texttt{. .}} \& \tiny{\texttt{<\symbol{92} .}}  \& \tiny{\texttt{< .}} \& \tiny{\texttt{<\symbol{92}\symbol{92} /\textsuperscript{*}}} \& \tiny{\texttt{. >\textsuperscript{*}}} \\
        \tiny{\texttt{ah\textsuperscript{tb}}} \& \tiny{\texttt{SH\_LA}} \&  \tiny{\texttt{SH}} \& \tiny{\texttt{SH\_LA\_LA}}  \& \tiny{\texttt{SH\_RA}}  \& \tiny{\texttt{SH}} \\
        \tiny{\texttt{c\textsuperscript{tb}}} \& \tiny{\texttt{SH}} \& \tiny{\texttt{SH\_LA}}  \& \tiny{\texttt{SH}} \& \tiny{\texttt{SH\_LA\_LA}} \& \tiny{\texttt{SH\_NOARC}} \\
                         \&                    \&                         \&                    \& \tiny{\texttt{\_NOARC\_RA}} \& \tiny{\texttt{\_RA}}\\
    \end{deptext}
    \depedge{3}{2}{}
    \depedge{5}{3}{}
    \depedge{5}{4}{}
    \depedge{1}{5}{}
    \depedge{4}{6}{}
\end{dependency}
\caption{\label{figure-example} Example of the linearizations used in this work in a sentence from the Afrikaans\textsubscript{AfriBooms} treebank. Dependency types are omitted for simplicity.}
\end{figure}

Sequence labeling assigns \emph{one} output label to every input token. Many problems are cast as sequence labeling due to its fast and simple nature, like PoS tagging, chunking, super tagging, named-entity recognition, semantic role labeling, and parsing. For dependency parsing, to create a linearized tree it suffices to assign each word $w_i$ a \emph{discrete} label of the form $(x_i, l_i)$, where $l_i$ is the dependency type and $x_i$ encodes a subset of the arcs of the tree related to such word. Although only labels seen in the training data can be predicted, \citet{strzyz-etal-2020-bracketing} show that the coverage is almost complete. We distinguish three families of encodings, which we now review (see also Figure \ref{figure-example}).

\paragraph{Head-selection encodings} \cite{spoustova2010dependency, li-etal-2018-seq2seq,strzyz-etal-2019-viable}. Each word label component $x_i$ encodes its head as an index or an (abstracted) offset. This can be done by labeling the target word with the (absolute) index of its head token, or by using a relative offset that accounts for the difference between the dependent and head indexes. In this work, we chose a relative PoS-based encoding (\texttt{rp\textsuperscript{h}}) that has shown to perform consistently better among the linerarizations of this family. Here, $x_i$ is a tuple $(p_i,o_i)$, such that if $o_i>0$ the head of $w_i$ is the $o_i$th word to the right of $w_i$ whose PoS tag is $p_i$; if $o_i<0$, the head of $w_i$ is the $o_i$th word to the left of $w_i$ whose PoS tag is $p_i$. Among its advantages, we find the capacity to encode any non-projective tree and words being directly and only linked to its head, but on the other hand it is dependent on external factors (e.g. PoS tags).\footnote{Other head-selection variants encode arcs based on word properties different than PoS tags \cite{lacroix2019dependency}.}
    
\paragraph{Bracketing-based encodings} \cite{yli-jyra-gomez-rodriguez-2017-generic}.
    Each $x_i$ encodes a sort of incoming and outgoing arcs of a given word and its neighbors, represented as bracket strings. More particularly, in \citet{strzyz-etal-2019-viable} each $x_i$ is a string that follows the expression \texttt{(<)?((\symbol{92})*|(/)*)(>)?}, where \texttt{<} means that $w_{i-1}$ has an incoming arc from the right, $k$ times \texttt{\symbol{92}} means that $w_i$ has $k$ outgoing arcs towards the left, $k$ times \texttt{/} means that $w_{i-1}$ has $k$ outgoing arcs to the right, and \texttt{>} means that $w_i$ has an arc coming from the left. This encoding produces a compressed label set while not relying on external features, such as PoS tags.
    However, when it comes to non-projectivity, it can only analyze crossing arcs in opposite directions. To counteract this, it is possible to define a linearization using a second independent pair of brackets (denoted with `*') to encode a 2-planar tree \cite{strzyz-etal-2020-bracketing}.\footnote{An $x$-planar tree can be separated into $x$ planes, where the arcs belonging to the same plane do not cross.}  In this work we are considering experiments with both the restricted non-projective (\texttt{rx\textsuperscript{b}}) and the 2-planar bracketing encodings (\texttt{2p\textsuperscript{b}}).
    
\paragraph{Transition-based encodings} \cite{gomez-rodriguez-etal-2020-unifying}. Each $x_i$ encodes a sub-sequence of the transitions to be generated by a \emph{left-to-right} transition-based parser. Given a sequence of transitions $t = t_1, ..., t_m $ with exactly $n$ read transitions\footnote{In left-to-right parsers, a read transition is an action that puts a word from the buffer into the stack. For algorithms such as the arc-standard or arc-hybrid this is only the \texttt{shift} action, while in the arc-eager both the \texttt{shift} and \texttt{right-arc} actions are read transitions. See also \cite{nivre2008algorithms}.}, it splits $t$ into $n$ chunks and assigns the $i$th chunk to the $i$th word. Its main advantage is more abstract, allowing to \emph{automatically derive encodings} relying on any left-to-right transition based parser (including dependency, constituency and semantic parsers). According to \citet{gomez-rodriguez-etal-2020-unifying}, they produce worse results than the bracketing encodings, but we include them in this work for completeness. In particular, we consider mappings from arc-hybrid \cite{kuhlmann-etal-2011-dynamic} (\texttt{ah\textsuperscript{tb}}) and \citet{covington2001fundamental} (\texttt{c\textsuperscript{tb}}), which are projective and non-projective transition-based algorithms.\\

\noindent To post-process corrupted predicted labels, we follow the heuristics described in each encoding paper.

\subsection{Sequence labeling framework}\label{section-framework}

\paragraph{Notes} Let $w$ be a sequence of words $[w_1,w_2,...,w_{|w|}]$, then $\vec{w}$ is a sequence of word vectors  that will be used as the input to our models. Each $\vec{w}_i$ will be a concatenation of: (i) a word embedding, (ii) a second word embedding computed through a char-LSTM, (iii) and \emph{optionally} a PoS tag embedding (we will discuss more about this last point in \S \ref{section-methodology-experiments}).\\

\noindent We use bidirectional long short-term memory networks \citep[biLSTMs;][]{hochreiter1997long,schuster1997bidirectional} to train our sequence labeling parsers. BiLSTMs are a strong baseline used in recent work across a number of tasks \cite{yang-zhang-2018-ncrf,reimers-gurevych-2017-reporting}. More particularly, we use two layers of biLSTMs, and each hidden vector $\vec{h_i}$ from the last biLSTM layer (associated to each input vector $\vec{w_i}$) is fed to separate feed-forward networks that are in charge of predicting each of the label components of the linearization (i.e. $x_i$ and $l_i$) using softmaxes, relying on hard-sharing multi-task learning
\citep[MTL;][]{caruana1997multitask,ruder2017overview}. 
Following \S \ref{section-encodings}, for all the encodings, except the 2-planar encoding, we will use a 2-task MTL setup: one task will predict $x_i$ according to each encoding specifics, and the other one will predict the dependency type, $l_i$. For the 2-planar bracketing encoding, which uses a second pair of brackets to predict the arcs from the second plane, we use instead a 3-task MTL setup, where the difference is that the prediction of $x_i$ is split into two tasks: one that predicts the first plane brackets and another task that predicts the brackets from the second plane.\footnote{We used 3 tasks because it establishes a more fair comparison in terms of label sparsity and follows previous work.}

It is worth noting that for this particular work we skipped computational expensive models, such as BERT \cite{devlin-etal-2019-bert}. There are three main reasons for this. First, the experiments in this paper imply training a total of 760 parsing models (see more details in \S \ref{section-methodology-experiments}), making the training on BERT (or variants) less practical. Second, there is not a multilingual or specific-language BERT model for all languages, and this could be the source of uncontrolled variables that could have an impact on the performance, and thereof on the conclusions.\footnote{Also, for the case of multilingual models, there is literature that concludes different about what makes a language beneficial for other under a BERT-based framework. For instance, \citet{wu-dredze-2019-beto} conclude that sharing a large amount sub-word pieces is important, while authors such as \citet{pires-etal-2019-multilingual} or \citet{artetxe-etal-2020-cross} state otherwise.} Third, even under the assumption of all language-specific BERT models being available, these are pre-trained on different data that add extra noise, which could be undesirable for our purpose.

\section{Methodology and experiments}\label{section-methodology-experiments}

We design two studies, detailed in \S \ref{ssec:experiment1} and \ref{ssec:experiment2}: 

\begin{enumerate}
    \item \label{enum-exp-1} We explore if some encodings are more data-efficient than others. To do so, we will simulate data-restricted setups, selecting rich-resource languages and using partial data. The goal is to test if some encodings are learnable with fewer data, or if other ones could obtain a better performance instead, but only under the assumption of very large data being available.
    
    \item \label{enum-exp-2} We focus on truly low-resource setups. This can be seen as a confirmation experiment to see if the findings under data-restricted setups hold for under-studied languages, and to confirm what sequence labeling linearizations are more recommendable under these conditions.
\end{enumerate}

\paragraph{Experimental setups} For  experiments \ref{enum-exp-1} and \ref{enum-exp-2}, we consider three setups that might have a different impact across the encodings:
\begin{enumerate}
\item \label{enum-setup-gold} Gold PoS tags setup: We train and run the models under an ideal framework that uses gold PoS tags as part of the input. The reason is that encodings such as \texttt{rp\textsuperscript{h}} rely on PoS tags to rebuild the linearized tree. This way, using gold PoS tags helps estimate the \emph{optimal} data-efficiency and learnability of these parsers under perfect (but unreal) conditions.
\item Predicted PoS tags setup: Setup \ref{enum-setup-gold} cannot truly reflect the performance that the encodings would obtain under real-world data-restricted conditions. Predicted PoS tags will be less helpful because their quality will degrade. This issue can affect more to the \texttt{rp\textsuperscript{h}} encoding, since it requires them to rebuild the tree from the labels, and miss-predicted PoS tags could propagate errors during decoding. Here, we train taggers for each treebank, using the same architecture used for the parsers. To be coherent with the data-restricted setups, taggers will be trained on the same amount of data used for the parsers. \ref{appendix:taggers-accuracy} discusses the PoS taggers performance.

\item No PoS tags setup: 
We train the models without using any PoS tags as part of the input.  It is worth noting that the setup is somewhat forced for the \texttt{rp\textsuperscript{h}} encoding, since we will still need to externally
run the taggers to obtain the PoS tags and rebuild the tree. Yet, we include the PoS-based encoding for completeness, and to have a better understanding about how different families of encodings suffer from not (or minimally) using PoS tags. For instance, that is a simple way to obtain simpler and faster parsing models, as part of the pipeline does not need to be executed, and the input vectors to the models will be smaller, translating into faster executions too. Also, in low-resource setups, PoS tags might not be available or the tagging models are not accurate enough to help deep learning models \cite{zhou2020pos,anderson2021taggers}.

\end{enumerate}

\subsection{Experiment 1: Encodings data-efficiency}
\label{ssec:experiment1}

\paragraph{Data} We chose 11 treebanks from UD2.7 \cite{ud2.7} with more than 10\,000 training sentences: German\textsubscript{HDT}, Czech\textsubscript{PDT}, Russian\textsubscript{SynTagRus}, Classical Chinese\textsubscript{Kyoto}, Persian\textsubscript{PDT}, Estonian\textsubscript{EDT}, Romanian\textsubscript{Nonstandard}, Korean\textsubscript{Kaist}, Ancient Greek\textsubscript{PROIEL}, Hindi\textsubscript{HDTB} and Latvian\textsubscript{LVTB}. They consider different families, scripts and levels of non-projectivity (see \ref{appendix:treebanks-info}). To simulate data-restricted setups, we created
training subsets of 100, 500, 1\,000, 5\,000 and 10\,000 samples, as well as the total training set. The training sets were shuffled before the division.

\paragraph{Setup} To assess the data-efficiency, we proceed as follows. As the  \texttt{rp\textsuperscript{h}} encoding has showed the strongest performance in previous work for multi-lingual setups \cite{strzyz-etal-2019-viable,gomez-rodriguez-etal-2020-unifying}, we are taking these models as the reference and an \emph{a priori} upper bound. Then, we compute the difference of the mean UAS (across the 11 treebanks) between the \texttt{rp\textsuperscript{h}} and each of the other linearizations, for all the models trained up to 10\,000 sentences. The goal is to determine which encodings suffer more when training with limited data and monitor to what extent the tendency holds as more data is introduced.
We compute the statistically significant difference between the \texttt{rp\textsuperscript{h}} and  the other encodings, using the p-value ($p<0.05$) of a paired t-test on the scores distribution, following recommended practices for dependency parsing \cite{dror-etal-2018-hitchhikers}. Finally, we show specific results for the models trained on the whole treebanks. In this work, we will report UAS over LAS, since the differences in the encodings lie in how they encode the dependency arcs and not their types.

\paragraph{Results} Tables \ref{table:comparison-gold-gold}, \ref{table:comparison-gold-predicted} and \ref{table:comparison-none} show the difference of the mean UAS for each encoding with respect to the \texttt{rp\textsuperscript{h}} one; for the gold PoS tags, predicted PoS tags and no PoS tags setups, respectively. For the gold PoS tags setup, the \texttt{rp\textsuperscript{h}} encoding performs better than the bracketing (\texttt{rx\textsuperscript{b}} and \texttt{2p\textsuperscript{b}}) and the transition-based (\texttt{ah\textsuperscript{tb}} and \texttt{c\textsuperscript{tb}}) encodings, for all the training splits. Yet, the gap narrows as the number of training sentence increases. For the predicted PoS tags setup, the relative PoS-based encoding  performs better for the smallest set of 100 sentences, but slightly worse for the sets of 500 and 1\,000 sentences with respect to \texttt{rx\textsuperscript{b}} and \texttt{2p\textsuperscript{b}}. With more data, the tendency resembles the one from the gold PoS tags setup. Third, for the setup without PoS tags, the tendency reverses. The bracketing encodings perform better, particularly for the smallest test sets, but the gap narrows as the number of training sentences increases.

\begin{table}[htpb]
\centering
\small{
\begin{tabular}{l|ccccc}
\# Sentences&\texttt{rp\textsuperscript{h}}&\texttt{rx\textsuperscript{b}}&\texttt{2p\textsuperscript{b}}&\texttt{ah\textsuperscript{tb}}&\texttt{c\textsuperscript{tb}}\\
\hline
100         &  68.34 &\cellcolor[rgb]{1,1,0.65}-2.15&\cellcolor[rgb]{1,1,0.62}-2.42&\cellcolor[rgb]{1,1,0.41}-5.82&\cellcolor[rgb]{1,1,0.29}-9.96\\
500         &  76.94 &\cellcolor[rgb]{1,1,0.72}-1.58&\cellcolor[rgb]{1,1,0.73}-1.5&\cellcolor[rgb]{1,1,0.43}-5.21&\cellcolor[rgb]{1,1,0.3}-9.35\\
1\,000        &  80.29 &\cellcolor[rgb]{1,1,0.74}-1.42&\cellcolor[rgb]{1,1,0.74}-1.43&\cellcolor[rgb]{1,1,0.44}-5.16&\cellcolor[rgb]{1,1,0.31}-8.9\\
5\,000        &  86.54 &\cellcolor[rgb]{1,1,0.78}-1.16&\cellcolor[rgb]{1,1,0.76}-1.26&\cellcolor[rgb]{1,1,0.53}-3.62&\cellcolor[rgb]{1,1,0.36}-7.04\\
10\,000       &  88.26 &\cellcolor[rgb]{1,1,0.83}-0.8&\cellcolor[rgb]{1,1,0.85}-0.72&\cellcolor[rgb]{1,1,0.53}-3.52&\cellcolor[rgb]{1,1,0.41}-5.67\\
\end{tabular}
}
\caption{Average UAS difference for the subsets of the rich-resource treebanks under the gold PoS tags setup. Blue and yellow cells show the UAS increase and decrease with respect to the \texttt{rp\textsuperscript{h}} encoding, respectively.}
\label{table:comparison-gold-gold}
\end{table}

\begin{table}[htpb]
\centering
\small{
\begin{tabular}{l|ccccc}
\# Sentences&\texttt{rp\textsuperscript{h}}&\texttt{rx\textsuperscript{b}}&\texttt{2p\textsuperscript{b}}&\texttt{ah\textsuperscript{tb}}&\texttt{c\textsuperscript{tb}}\\
\hline
100         &  41.87 &\cellcolor[rgb]{1,1,0.9}-0.42&\cellcolor[rgb]{1,1,0.96}-0.19&\cellcolor[rgb]{1,1,0.68}-1.9&\cellcolor[rgb]{1,1,0.53}-3.59\\
500         &  63.45 &\cellcolor[rgb]{1,1,1.0}-0.01&\cellcolor[rgb]{0.97,0.97,1}0.14&\cellcolor[rgb]{1,1,0.67}-1.96&\cellcolor[rgb]{1,1,0.41}-5.73\\
1\.000        &  68.10 &\cellcolor[rgb]{0.94,0.94,1}0.25&\cellcolor[rgb]{0.96,0.96,1}0.17&\cellcolor[rgb]{1,1,0.62}-2.44&\cellcolor[rgb]{1,1,0.42}-5.53\\
5\,000        &  78.56 &\cellcolor[rgb]{1,1,0.87}-0.62&\cellcolor[rgb]{1,1,0.86}-0.63&\cellcolor[rgb]{1,1,0.61}-2.53&\cellcolor[rgb]{1,1,0.42}-5.44\\
10\,000       &  82.29 &\cellcolor[rgb]{1,1,0.91}-0.37&\cellcolor[rgb]{1,1,0.92}-0.36&\cellcolor[rgb]{1,1,0.62}-2.49&\cellcolor[rgb]{1,1,0.47}-4.44\\
\end{tabular}
}
\caption{Average UAS difference for the subsets of the rich-resource treebanks under the predicted PoS tags setup.}
\label{table:comparison-gold-predicted}
\end{table}

\begin{table}[!htpb]
\centering
\small{
\begin{tabular}{l|ccccc}
\# Sentences&\texttt{rp\textsuperscript{h}}&\texttt{rx\textsuperscript{b}}&\texttt{2p\textsuperscript{b}}&\texttt{ah\textsuperscript{tb}}&\texttt{c\textsuperscript{tb}}\\
\hline
100         &  35.60 &\cellcolor[rgb]{0.31,0.31,1}9.06&\cellcolor[rgb]{0.3,0.3,1}9.31&\cellcolor[rgb]{0.35,0.35,1}7.57&\cellcolor[rgb]{0.45,0.45,1}4.83\\
500         &  58.63 &\cellcolor[rgb]{0.57,0.57,1}3.04&\cellcolor[rgb]{0.62,0.62,1}2.45&\cellcolor[rgb]{0.8,0.8,1}0.99&\cellcolor[rgb]{1,1,0.64}-2.26\\
1\,000        &  63.99 &\cellcolor[rgb]{0.53,0.53,1}3.59&\cellcolor[rgb]{0.54,0.54,1}3.42&\cellcolor[rgb]{0.83,0.83,1}0.83&\cellcolor[rgb]{1,1,0.64}-2.24\\
5\,000        &  75.57 &\cellcolor[rgb]{0.73,0.73,1}1.47&\cellcolor[rgb]{0.72,0.72,1}1.55&\cellcolor[rgb]{1,1,0.96}-0.19&\cellcolor[rgb]{1,1,0.57}-3.07\\
10\,000       &  79.90 &\cellcolor[rgb]{0.77,0.77,1}1.22&\cellcolor[rgb]{0.72,0.72,1}1.54&\cellcolor[rgb]{1,1,0.82}-0.87&\cellcolor[rgb]{1,1,0.58}-2.93\\
\end{tabular}
}
\caption{Average UAS difference for the subsets of the rich-resource treebanks under the no PoS tags setup.}
\label{table:comparison-none}
\end{table}

\paragraph{Discussion} 
The results from the experiments shed light about differences existing across different encodings and running configurations.
First, under an ideal, gold environment, the \texttt{rp\textsuperscript{h}} encoding makes a better use of limited data than the bracketing and transition-based encodings. Second, the predicted PoS tag setup shows that the performance of the PoS taggers can have a significant impact on the performance for the \texttt{rp\textsuperscript{h}} encoding. More interestingly, weaknesses from different encodings seem to manifest to different extents depending on the amount of training data. For instance,  when training data is scarce (100 sentences), bracketing encodings still cannot
outperform the \texttt{rp\textsuperscript{h}} encoding, despite the lower performance of the PoS taggers. However, when working with setups ranging from 500 to 1000 sentences, 
there is a slight advantage of the bracketing encodings with respect to \texttt{rp\textsuperscript{h}}, suggesting that with this amount of data, bracketing encodings could be the preferable choice, since they seem able to exploit their potential in a better way than the \texttt{rp\textsuperscript{h}} encoding can exploit not fully accurate PoS tags. With more training samples, the relative PoS-based encoding is again the best performing model across the board. In \S \ref{ssec:experiment2} we will discuss deeper how for truly low-resources languages the advantage in favour of bracketing representations exacerbates more for the predicted and no PoS tags setups.

\begin{table}[htpb]
\begin{center}
\small{
\begin{tabular}{l|ccccc}
&\texttt{rp\textsuperscript{h}}&\texttt{rx\textsuperscript{b}}&\texttt{2p\textsuperscript{b}}&\texttt{ah\textsuperscript{tb}}&\texttt{c\textsuperscript{tb}}\\
\hline
grc&$83.09$&\cellcolor[HTML]{ff6e6e} $79.89^{ -- }$&\cellcolor[HTML]{ffff6f} $81.7^{ - }$&\cellcolor[HTML]{ff6e6e} $78.57^{ -- }$&\cellcolor[HTML]{ff6e6e} $79.86^{ -- }$\\
lzh&$90.21$&\cellcolor[HTML]{ffff6f} $89.56^{ - }$&\cellcolor[HTML]{ffff6f} $89.24^{ - }$&\cellcolor[HTML]{ff6e6e} $89.04^{ -- }$&\cellcolor[HTML]{ffff6f} $89.18^{ - }$\\
cs&$91.61$&\cellcolor[HTML]{ff6e6e} $90.49^{ -- }$&\cellcolor[HTML]{ff6e6e} $90.91^{ -- }$&\cellcolor[HTML]{ff6e6e} $88.18^{ -- }$&\cellcolor[HTML]{ff6e6e} $85.64^{ -- }$\\
et&$85.62$&\cellcolor[HTML]{ffff6f} $84.79^{ - }$&\cellcolor[HTML]{ffff6f} $84.91^{ - }$&\cellcolor[HTML]{ff6e6e} $81.86^{ -- }$&\cellcolor[HTML]{ff6e6e} $81.11^{ -- }$\\
de&$96.69$&\cellcolor[HTML]{ff6e6e} $95.95^{ -- }$&\cellcolor[HTML]{ff6e6e} $96.38^{ -- }$&\cellcolor[HTML]{ff6e6e} $95.15^{ -- }$&\cellcolor[HTML]{ff6e6e} $86.51^{ -- }$\\
hi&$94.69$&\cellcolor[HTML]{ff6e6e} $94.09^{ -- }$&\cellcolor[HTML]{ffff6f} $94.43^{ - }$&\cellcolor[HTML]{ff6e6e} $93.05^{ -- }$&\cellcolor[HTML]{ff6e6e} $85.02^{ -- }$\\
ko&$87.26$&\cellcolor[HTML]{ff6e6e} $86.24^{ -- }$&\cellcolor[HTML]{ffff6f} $86.52^{ - }$&\cellcolor[HTML]{ff6e6e} $85.68^{ -- }$&\cellcolor[HTML]{ff6e6e} $84.06^{ -- }$\\
lv&$85.3$&\cellcolor[HTML]{ffff6f} $83.88^{ - }$&\cellcolor[HTML]{ffff6f} $84.01^{ - }$&\cellcolor[HTML]{ff6e6e} $80.88^{ -- }$&\cellcolor[HTML]{ff6e6e} $81.38^{ -- }$\\
fa&$92.61$&\cellcolor[HTML]{ffff6f} $92.07^{ - }$&\cellcolor[HTML]{ffff6f} $92.44^{ - }$&\cellcolor[HTML]{ff6e6e} $90.45^{ -- }$&\cellcolor[HTML]{ff6e6e} $87.09^{ -- }$\\
ro&$90.49$&\cellcolor[HTML]{ffff6f} $89.68^{ - }$&\cellcolor[HTML]{ff6e6e} $89.63^{ -- }$&\cellcolor[HTML]{ff6e6e} $87.39^{ -- }$&\cellcolor[HTML]{ff6e6e} $86.38^{ -- }$\\
ru&$91.23$&\cellcolor[HTML]{ff6e6e} $90.1^{ -- }$&\cellcolor[HTML]{ff6e6e} $90.1^{ -- }$&\cellcolor[HTML]{ff6e6e} $88.19^{ -- }$&\cellcolor[HTML]{ff6e6e} $84.96^{ -- }$\\
\hline 
Avg & \textbf{89.89} & 88.79 & 89.12 & 87.13 & 84.65
\end{tabular}
}
\caption{UAS for the rich-resource treebanks, using the whole training set and the gold PoS tags setup. The red ({\small -\,-}) and green cells ({\small ++}) show that a given encoding performed worse or better than the \texttt{rp\textsuperscript{h}} model, and that the difference is statistically significant. Lime and yellow cells mean that there is no a significant difference between a given encoding and the \texttt{rp\textsuperscript{h}}, appending a \textsuperscript{$+$} or a \textsuperscript{$-$} when they performed better or worse than the \texttt{rp\textsuperscript{h}}.}
\label{table:hr-gold-gold}
\end{center}
\end{table}

\begin{table}[htpb]
\begin{center}
\small{
\begin{tabular}{l|ccccc}
&\texttt{rp\textsuperscript{h}}&\texttt{rx\textsuperscript{b}}&\texttt{2p\textsuperscript{b}}&\texttt{ah\textsuperscript{tb}}&\texttt{c\textsuperscript{tb}}\\
\hline
grc&$80.2$&\cellcolor[HTML]{ff6e6e} $77.61^{ -- }$&\cellcolor[HTML]{ffff6f} $79.21^{ - }$&\cellcolor[HTML]{ff6e6e} $76.49^{ -- }$&\cellcolor[HTML]{ffff6f} $77.71^{ - }$\\
lzh&$79.93$&\cellcolor[HTML]{ffff6f} $79.8^{ - }$&\cellcolor[HTML]{ffff6f} $79.42^{ - }$&\cellcolor[HTML]{ffff6f} $79.41^{ - }$&\cellcolor[HTML]{ffff6f} $79.54^{ - }$\\
cs&$90.04$&\cellcolor[HTML]{ff6e6e} $88.93^{ -- }$&\cellcolor[HTML]{ff6e6e} $89.34^{ -- }$&\cellcolor[HTML]{ff6e6e} $86.67^{ -- }$&\cellcolor[HTML]{ff6e6e} $84.25^{ -- }$\\
et&$81.07$&\cellcolor[HTML]{ffff6f} $80.36^{ - }$&\cellcolor[HTML]{ffff6f} $80.34^{ - }$&\cellcolor[HTML]{ff6e6e} $77.71^{ -- }$&\cellcolor[HTML]{ff6e6e} $76.95^{ -- }$\\
de&$95.85$&\cellcolor[HTML]{ff6e6e} $95.14^{ -- }$&\cellcolor[HTML]{ff6e6e} $95.54^{ -- }$&\cellcolor[HTML]{ff6e6e} $94.34^{ -- }$&\cellcolor[HTML]{ff6e6e} $85.79^{ -- }$\\
hi&$92.22$&\cellcolor[HTML]{ffff6f} $91.76^{ - }$&\cellcolor[HTML]{ffff6f} $92.21^{ - }$&\cellcolor[HTML]{ff6e6e} $90.72^{ -- }$&\cellcolor[HTML]{ff6e6e} $83.24^{ -- }$\\
ko&$84.25$&\cellcolor[HTML]{ffff6f} $83.44^{ - }$&\cellcolor[HTML]{ffff6f} $83.42^{ - }$&\cellcolor[HTML]{ff6e6e} $82.98^{ -- }$&\cellcolor[HTML]{ff6e6e} $81.25^{ -- }$\\
lv&$70.65$&\cellcolor[HTML]{6fff6f} $71.98^{ ++ }$&\cellcolor[HTML]{deffbc} $71.08^{ + }$&\cellcolor[HTML]{ffff6f} $68.9^{ - }$&\cellcolor[HTML]{ffff6f} $68.97^{ - }$\\
fa&$90.39$&\cellcolor[HTML]{ffff6f} $89.8^{ - }$&\cellcolor[HTML]{ffff6f} $90.32^{ - }$&\cellcolor[HTML]{ff6e6e} $88.27^{ -- }$&\cellcolor[HTML]{ff6e6e} $85.28^{ -- }$\\
ro&$87.32$&\cellcolor[HTML]{ffff6f} $86.64^{ - }$&\cellcolor[HTML]{ffff6f} $86.49^{ - }$&\cellcolor[HTML]{ff6e6e} $84.44^{ -- }$&\cellcolor[HTML]{ff6e6e} $83.5^{ -- }$\\
ru&$88.71$&\cellcolor[HTML]{ffff6f} $88.13^{ - }$&\cellcolor[HTML]{ffff6f} $88.24^{ - }$&\cellcolor[HTML]{ff6e6e} $85.93^{ -- }$&\cellcolor[HTML]{ff6e6e} $82.96^{ -- }$\\
\hline 
Avg & \textbf{85.51} & 84.87 & 85.06 & 83.26 & 80.86
\end{tabular}
}
\caption{UAS for the rich-resource treebanks, using the whole training set and the predicted PoS tags setup.}
\label{table:hr-gold-predicted}

\end{center}
\end{table}

\begin{table}[htpb]
\begin{center}
\small{
\begin{tabular}{l|ccccc}
&\texttt{rp\textsuperscript{h}}&\texttt{rx\textsuperscript{b}}&\texttt{2p\textsuperscript{b}}&\texttt{ah\textsuperscript{tb}}&\texttt{c\textsuperscript{tb}}\\
\hline
grc&$77.84$&\cellcolor[HTML]{ffff6f} $77.41^{ - }$&\cellcolor[HTML]{deffbc} $79.16^{ + }$&\cellcolor[HTML]{ffff6f} $75.64^{ - }$&\cellcolor[HTML]{ffff6f} $76.99^{ - }$\\
lzh&$79.99$&\cellcolor[HTML]{deffbc} $81.02^{ + }$&\cellcolor[HTML]{deffbc} $80.75^{ + }$&\cellcolor[HTML]{6fff6f} $81.11^{ ++ }$&\cellcolor[HTML]{6fff6f} $81.42^{ ++ }$\\
cs&$88.67$&\cellcolor[HTML]{ffff6f} $88.2^{ - }$&\cellcolor[HTML]{ffff6f} $88.64^{ - }$&\cellcolor[HTML]{ff6e6e} $85.8^{ -- }$&\cellcolor[HTML]{ff6e6e} $84.23^{ -- }$\\
et&$77.85$&\cellcolor[HTML]{6fff6f} $79.69^{ ++ }$&\cellcolor[HTML]{6fff6f} $79.99^{ ++ }$&\cellcolor[HTML]{ffff6f} $77.15^{ - }$&\cellcolor[HTML]{ffff6f} $76.27^{ - }$\\
de&$94.51$&\cellcolor[HTML]{6fff6f} $95.09^{ ++ }$&\cellcolor[HTML]{6fff6f} $95.41^{ ++ }$&\cellcolor[HTML]{ffff6f} $94.18^{ - }$&\cellcolor[HTML]{ff6e6e} $83.54^{ -- }$\\
hi&$89.43$&\cellcolor[HTML]{6fff6f} $91.7^{ ++ }$&\cellcolor[HTML]{6fff6f} $91.98^{ ++ }$&\cellcolor[HTML]{6fff6f} $90.72^{ ++ }$&\cellcolor[HTML]{ff6e6e} $82.86^{ -- }$\\
ko&$79.39$&\cellcolor[HTML]{6fff6f} $82.18^{ ++ }$&\cellcolor[HTML]{6fff6f} $82.15^{ ++ }$&\cellcolor[HTML]{6fff6f} $81.88^{ ++ }$&\cellcolor[HTML]{6fff6f} $80.3^{ ++ }$\\
lv&$62.56$&\cellcolor[HTML]{6fff6f} $71.17^{ ++ }$&\cellcolor[HTML]{6fff6f} $72.38^{ ++ }$&\cellcolor[HTML]{6fff6f} $66.78^{ ++ }$&\cellcolor[HTML]{6fff6f} $69.38^{ ++ }$\\
fa&$89.14$&\cellcolor[HTML]{6fff6f} $90.39^{ ++ }$&\cellcolor[HTML]{6fff6f} $90.48^{ ++ }$&\cellcolor[HTML]{ffff6f} $88.49^{ - }$&\cellcolor[HTML]{ff6e6e} $84.54^{ -- }$\\
ro&$85.28$&\cellcolor[HTML]{deffbc} $86.41^{ + }$&\cellcolor[HTML]{6fff6f} $86.94^{ ++ }$&\cellcolor[HTML]{ffff6f} $84.25^{ - }$&\cellcolor[HTML]{ff6e6e} $83.04^{ -- }$\\
ru&$83.35$&\cellcolor[HTML]{6fff6f} $83.98^{ ++ }$&\cellcolor[HTML]{6fff6f} $84.5^{ ++ }$&\cellcolor[HTML]{6fff6f} $83.42^{ ++ }$&\cellcolor[HTML]{ff6e6e} $80.26^{ -- }$\\
\hline 
Avg & 82.55 & 84.29 & \textbf{84.76} & 82.67 & 80.26
\end{tabular}
}
\caption{UAS for the rich-resource treebanks, using the whole training set and the no PoS tags setup.}
\label{table:hr-none}
\end{center}
\end{table}

Tables \ref{table:hr-gold-gold}, \ref{table:hr-gold-predicted} and \ref{table:hr-none} show the UAS on the full training sets of the rich-resource treebanks for the gold PoS tags, predicted PoS tags, and no PoS tags setups. The goal is to show if under large amounts of data some of the encodings could perform on par with \texttt{rp\textsuperscript{h}}, since Tables \ref{table:comparison-gold-gold} and \ref{table:comparison-gold-predicted} indicated that differences in performance across encodings decreased when the number of training samples increase. Although performance across encodings becomes closer, their ranking remains the same.

\subsection{Experiment 2: Encodings performance on truly low-resource languages}
\label{ssec:experiment2}

\paragraph{Data} We choose the 10 smallest treebanks\footnote{Code switching treebanks and small treebanks of rich-resource languages were not considered.} (in terms of training sentences) that had a dev set: Lithuanian\textsubscript{HSE}, Marathi\textsubscript{UFAL}, Hungarian\textsubscript{Szeged}, Telugu\textsubscript{MTG}, Tamil\textsubscript{TTB}, Faroese\textsubscript{FarPaHC}, Coptic\textsubscript{Scriptorium}, Maltese\textsubscript{MUDT}, Wolof\textsubscript{WTB} and Afrikaans\textsubscript{AfriBooms} (see \ref{appendix:treebanks-info}). Their sizes range between 153 and 1350 training sentences, most being around or between 500 and 1\,000 
(see \ref{appendix:lr-treebanks-size}).

\paragraph{Setup} We rerun a subset of the experiments from \S \ref{ssec:experiment1}, to check if the results follow the same trends, and conclusions are therefore similar.

\begin{table}
\begin{center}
\small{
\begin{tabular}{l|ccccc}
&\texttt{rp\textsuperscript{h}}&\texttt{rx\textsuperscript{b}}&\texttt{2p\textsuperscript{b}}&\texttt{ah\textsuperscript{tb}}&\texttt{c\textsuperscript{tb}}\\
\hline
af&$88.02$&\cellcolor[HTML]{ff6e6e} $85.7^{ -- }$&\cellcolor[HTML]{ff6e6e} $85.48^{ -- }$&\cellcolor[HTML]{ff6e6e} $81.84^{ -- }$&\cellcolor[HTML]{ff6e6e} $78.6^{ -- }$\\
cop&$88.73$&\cellcolor[HTML]{ffff6f} $88.43^{ - }$&\cellcolor[HTML]{ffff6f} $88.72^{ - }$&\cellcolor[HTML]{ff6e6e} $85.5^{ -- }$&\cellcolor[HTML]{ff6e6e} $84.35^{ -- }$\\
fo&$84.04$&\cellcolor[HTML]{ffff6f} $83.76^{ - }$&\cellcolor[HTML]{deffbc} $84.09^{ + }$&\cellcolor[HTML]{ff6e6e} $81.78^{ -- }$&\cellcolor[HTML]{ff6e6e} $79.53^{ -- }$\\
hu&$79.75$&\cellcolor[HTML]{ff6e6e} $76.14^{ -- }$&\cellcolor[HTML]{ff6e6e} $76.13^{ -- }$&\cellcolor[HTML]{ff6e6e} $71.66^{ -- }$&\cellcolor[HTML]{ff6e6e} $64.27^{ -- }$\\
lt&$51.98$&\cellcolor[HTML]{ffff6f} $50.28^{ - }$&\cellcolor[HTML]{ffff6f} $50.19^{ - }$&\cellcolor[HTML]{ffff6f} $45.0^{ - }$&\cellcolor[HTML]{ffff6f} $46.6^{ - }$\\
mt&$81.81$&\cellcolor[HTML]{ffff6f} $81.05^{ - }$&\cellcolor[HTML]{ffff6f} $80.82^{ - }$&\cellcolor[HTML]{ff6e6e} $76.78^{ -- }$&\cellcolor[HTML]{ff6e6e} $74.98^{ -- }$\\
mr&$77.43$&\cellcolor[HTML]{ffff6f} $76.46^{ - }$&\cellcolor[HTML]{ffff6f} $75.97^{ - }$&\cellcolor[HTML]{ffff6f} $76.94^{ - }$&\cellcolor[HTML]{ffff6f} $73.54^{ - }$\\
ta&$74.96$&\cellcolor[HTML]{ffff6f} $73.1^{ - }$&\cellcolor[HTML]{ffff6f} $71.9^{ - }$&\cellcolor[HTML]{ffff6f} $71.74^{ - }$&\cellcolor[HTML]{ff6e6e} $66.01^{ -- }$\\
te&$90.01$&\cellcolor[HTML]{deffbc} $91.26^{ + }$&\cellcolor[HTML]{deffbc} $90.43^{ + }$&\cellcolor[HTML]{deffbc} $90.01^{ + }$&\cellcolor[HTML]{ffff6f} $89.46^{ - }$\\
wo&$86.19$&\cellcolor[HTML]{ff6e6e} $84.64^{ -- }$&\cellcolor[HTML]{ff6e6e} $84.51^{ -- }$&\cellcolor[HTML]{ff6e6e} $80.65^{ -- }$&\cellcolor[HTML]{ff6e6e} $77.43^{ -- }$\\
\hline 
Avg & \textbf{80.29} & 79.08 & 78.82 & 76.19 & 73.48
\end{tabular}
}
\end{center}
\caption{UAS for the low-resource treebanks for the gold PoS tags setup.}
\label{table:lr-gold-gold}
\end{table}

\begin{table}
\begin{center}
\small{
\begin{tabular}{l|ccccc}
&\texttt{rp\textsuperscript{h}}&\texttt{rx\textsuperscript{b}}&\texttt{2p\textsuperscript{b}}&\texttt{ah\textsuperscript{tb}}&\texttt{c\textsuperscript{tb}}\\
\hline
af&$81.84$&\cellcolor[HTML]{ffff6f} $80.29^{ - }$&\cellcolor[HTML]{ffff6f} $79.9^{ - }$&\cellcolor[HTML]{ff6e6e} $77.3^{ -- }$&\cellcolor[HTML]{ff6e6e} $73.61^{ -- }$\\
cop&$85.77$&\cellcolor[HTML]{deffbc} $86.25^{ + }$&\cellcolor[HTML]{deffbc} $85.92^{ + }$&\cellcolor[HTML]{ff6e6e} $83.14^{ -- }$&\cellcolor[HTML]{ff6e6e} $81.84^{ -- }$\\
fo&$77.04$&\cellcolor[HTML]{ffff6f} $76.97^{ - }$&\cellcolor[HTML]{deffbc} $77.52^{ + }$&\cellcolor[HTML]{ffff6f} $75.23^{ - }$&\cellcolor[HTML]{ffff6f} $74.24^{ - }$\\
hu&$70.52$&\cellcolor[HTML]{ffff6f} $68.51^{ - }$&\cellcolor[HTML]{ffff6f} $68.77^{ - }$&\cellcolor[HTML]{ff6e6e} $64.98^{ -- }$&\cellcolor[HTML]{ff6e6e} $58.37^{ -- }$\\
lt&$30.28$&\cellcolor[HTML]{deffbc} $34.53^{ + }$&\cellcolor[HTML]{deffbc} $33.11^{ + }$&\cellcolor[HTML]{deffbc} $31.23^{ + }$&\cellcolor[HTML]{ffff6f} $29.91^{ - }$\\
mt&$74.6$&\cellcolor[HTML]{deffbc} $75.64^{ + }$&\cellcolor[HTML]{deffbc} $75.07^{ + }$&\cellcolor[HTML]{ff6e6e} $71.17^{ -- }$&\cellcolor[HTML]{ff6e6e} $70.35^{ -- }$\\
mr&$66.99$&\cellcolor[HTML]{deffbc} $67.96^{ + }$&\cellcolor[HTML]{deffbc} $67.23^{ + }$&\cellcolor[HTML]{deffbc} $68.93^{ + }$&\cellcolor[HTML]{deffbc} $67.23^{ + }$\\
ta&$57.11$&\cellcolor[HTML]{deffbc} $60.73^{ + }$&\cellcolor[HTML]{deffbc} $57.57^{ + }$&\cellcolor[HTML]{deffbc} $58.77^{ + }$&\cellcolor[HTML]{ffff6f} $55.51^{ - }$\\
te&$86.41$&\cellcolor[HTML]{deffbc} $87.93^{ + }$&\cellcolor[HTML]{deffbc} $87.93^{ + }$&\cellcolor[HTML]{deffbc} $86.96^{ + }$&\cellcolor[HTML]{deffbc} $86.69^{ + }$\\
wo&$76.88$&\cellcolor[HTML]{ffff6f} $76.4^{ - }$&\cellcolor[HTML]{ffff6f} $76.3^{ - }$&\cellcolor[HTML]{ff6e6e} $73.24^{ -- }$&\cellcolor[HTML]{ff6e6e} $70.84^{ -- }$\\
\hline 
 Avg & 70.74 & \textbf{71.52} & 70.93 & 69.10 & 66.86
\end{tabular}
}
\caption{UAS for the low-resource treebanks for the predicted PoS tags setup.}
\label{table:lr-gold-predicted}
\end{center}
\end{table}

\begin{table}
\begin{center}
\small{
\begin{tabular}{l|ccccc}
&\texttt{rp\textsuperscript{h}}&\texttt{rx\textsuperscript{b}}&\texttt{2p\textsuperscript{b}}&\texttt{ah\textsuperscript{tb}}&\texttt{c\textsuperscript{tb}}\\
\hline
af&$79.86$&\cellcolor[HTML]{deffbc} $80.78^{ + }$&\cellcolor[HTML]{deffbc} $80.07^{ + }$&\cellcolor[HTML]{ff6e6e} $75.47^{ -- }$&\cellcolor[HTML]{ff6e6e} $73.76^{ -- }$\\
cop&$84.36$&\cellcolor[HTML]{deffbc} $85.76^{ + }$&\cellcolor[HTML]{deffbc} $85.13^{ + }$&\cellcolor[HTML]{ffff6f       } $83.07^{ - }$&\cellcolor[HTML]{ff6e6e} $81.28^{ -- }$\\
fo&$73.98$&\cellcolor[HTML]{6fff6f} $77.08^{ ++ }$&\cellcolor[HTML]{6fff6f} $77.04^{ ++ }$&\cellcolor[HTML]{deffbc} $75.07^{ + }$&\cellcolor[HTML]{ffff6f} $73.67^{ - }$\\
hu&$63.63$&\cellcolor[HTML]{deffbc} $65.21^{ + }$&\cellcolor[HTML]{deffbc} $64.8^{ + }$&\cellcolor[HTML]{ffff6f     } $62.04^{ - }$&\cellcolor[HTML]{ff6e6e} $56.17^{ -- }$\\
lt&$26.89$&\cellcolor[HTML]{6fff6f} $34.62^{ ++ }$&\cellcolor[HTML]{6fff6f} $35.38^{ ++ }$&\cellcolor[HTML]{6fff6f} $34.06^{ ++ }$&\cellcolor[HTML]{deffbc} $32.92^{ + }$\\
mt&$70.95$&\cellcolor[HTML]{6fff6f} $75.5^{ ++ }$&\cellcolor[HTML]{6fff6f} $75.3^{ ++ }$&\cellcolor[HTML]{deffbc} $71.69^{ + }$&\cellcolor[HTML]{ffff6f} $70.32^{ - }$\\
mr&$64.08$&\cellcolor[HTML]{deffbc} $66.75^{ + }$&\cellcolor[HTML]{deffbc} $67.96^{ + }$&\cellcolor[HTML]{deffbc} $69.66^{ + }$&\cellcolor[HTML]{deffbc} $64.56^{ + }$\\
ta&$52.79$&\cellcolor[HTML]{6fff6f} $60.03^{ ++ }$&\cellcolor[HTML]{deffbc} $56.61^{ + }$&\cellcolor[HTML]{6fff6f} $59.58^{ ++ }$&\cellcolor[HTML]{deffbc} $54.95^{ + }$\\
te&$85.44$&\cellcolor[HTML]{deffbc} $88.49^{ + }$&\cellcolor[HTML]{deffbc} $88.63^{ + }$&\cellcolor[HTML]{deffbc} $87.1^{ + }$&\cellcolor[HTML]{deffbc} $86.82^{ + }$\\
wo&$73.11$&\cellcolor[HTML]{6fff6f} $77.17^{ ++ }$&\cellcolor[HTML]{6fff6f} $76.95^{ ++ }$&\cellcolor[HTML]{deffbc} $74.01^{ + }$&\cellcolor[HTML]{ffff6f} $70.86^{ - }$\\
\hline 
Avg & 67.51 & \textbf{71.14} & 70.79 & 69.18 & 66.53
\end{tabular}
}
\caption{UAS for the low-resource treebanks for the no PoS tags setup.}
\label{table:lr-none}
\end{center}
\end{table}

\paragraph{Results} Tables \ref{table:lr-gold-gold}, \ref{table:lr-gold-predicted} and \ref{table:lr-none} show the UAS for each encoding and treebank for the gold PoS tags setup, the predicted PoS tags setup and the no PoS tags setup, respectively. Again, under perfect conditions, the relative PoS-based encoding performs overall better, except for Telugu, which seems to be an outlier. For the predicted PoS tags setup, the bracketing-based encodings perform consistently better for most of the treebanks. 
For the no PoS tags setup, the bracketing-based encodings obtain, on average, more than 3 points than the relative PoS-head selection encoding, which even performs worse than the transition-based encodings. 

\paragraph{Discussion} These experiments help elaborate on the findings of \S \ref{ssec:experiment1}.
With respect to the ideal gold PoS tags setup, things do not change much, and the relative PoS-based encoding performs overall better. Still, this should not be taken as a ground truth about how the encodings will perform in real-world setups. For instance, for the predicted PoS tags setup, the bracketing-based encodings perform consistently better in most of the treebanks. This reinforces some of the suspicions found in the experiments of Table \ref{table:comparison-gold-predicted}, where training on rich-resource languages, but with limited data, revealed that bracketing encodings performed better, although just slightly. Also, it is worth noting that most of the low-resource treebanks tested in this work have a number of training sentences in the range where the bracketing-based encodings performed better for the predicted PoS tags setup in Table \ref{table:comparison-gold-predicted}, i.e. from 500 to 1\,000 sentences (see \ref{appendix:lr-treebanks-size}).
Yet, the better performance of bracketing encodings is more evident when running on real low-resource treebanks. This does not only suggest that the bracketing encodings are better for real low-resource sequence labeling parsing, but it could also pose more general limitations for other low-resource NLP tasks that are evaluated only on `faked' low-resource setups, and that could lead to incomplete or even misleading conclusions.

Overall, the results suggest that bracketing encodings are the most suitable linearizations for real low-resource sequence labeling parsing.

\section{Conclusion}
We have studied sequence labeling encodings for dependency parsing in low-resource setups. First, we explored which encodings are more data-efficient under different conditions that include the use of gold PoS tags, predicted PoS tags and no PoS tags as part of the input. By restricting training data for rich-resource treebanks, we observe that although bracketing encodings are less data-efficient than head-selection ones under ideal conditions, this disadvantage can vanish when the input conditions are not gold and data is limited. Second, we studied their performance under the same running configurations, but on truly low-resource languages. These results show more clearly the greatest utility of bracketing encodings over the rest of the ones when training data is limited and the quality of external factors, such as PoS tags, is affected by the low-resource nature of the problem.

\section*{Acknowledgements}

This work is supported by a 2020 Leonardo Grant for Researchers and Cultural Creators from the FBBVA.\footnote{FBBVA accepts no responsibility for the opinions, statements and contents included in the project and/or the results thereof, which are entirely the responsibility of the authors.} The work also receives funding
from the European Research Council (FASTPARSE, grant agreement No 714150), from ERDF/MICINN-AEI (ANSWER-ASAP, TIN2017-85160-C2-1-R, SCANNER, PID2020-113230RB-C21), from Xunta de Galicia (ED431C 2020/11), and from Centro de Investigación de Galicia `CITIC', funded by Xunta de Galicia and the European Union (European Regional Development Fund- Galicia 2014-2020 Program) by grant ED431G 2019/01.
\bibliographystyle{acl_natbib}
\bibliography{anthology,ranlp2021}
%\clearpage

\vfill\eject
\appendix
\renewcommand{\thesection}{Appendix \Alph{section}}

\section{Taggers accuracy}

%\section{Appendices}
%\subsection{Taggers accuracy}
%\columnbreak
%\vspace{1cm}
%Table \ref{table:taggers-accuracy} shows the average accuracy of the rich-resource taggers using different amounts of training data.

\label{appendix:taggers-accuracy}
\begin{table}[!htpb]
\begin{center}
\small{
\begin{tabular}{l|c} 
\# Sentences & Average accuracy \\ \hline 100     &  57.08  \\500     &  81.24 \\1\,000    &  85.03 \\5\,000  &  90.90 \\10\,000   &  92.77\\
Low-resource        &  85.03
\end{tabular}
}
\caption{Average accuracy of the taggers for the splits of the rich-resource treebanks and the complete low-resource treebanks.}
\label{table:taggers-accuracy}
\end{center}
\end{table}

\section{Treebanks information}
\label{appendix:treebanks-info}

\begin{table}[hbtp]
\begin{center}
\small{
\begin{tabular}{l|cll}
 & \thead{\% Non-projective\\sentences} & Family & Script\\
\hline
de & 6.76 & IE (Germanic) & Latin\\
cs & 11.49 & IE (West Slavic) & Latin \\
ru & 7.53 & IE (East Slavic) & Cyrillic \\
lzh & 0.01 & Sino-Tibetan & Chinese characters\\
fa & 14.22 & IE (Iranian) & Persian\\
et & 3.22 & Uralic & Latin\\
ro & 5.43 & IE (Romance) & Latin\\
ko & 21.70 & Korean & Korean\\
grc& 37.52 & IE (Greek) & Greek\\
hi & 13.60 & IE (Indo-Aryan) & Devanagari\\
lv & 6.53 & IE (Baltic) & Latin\\
af & 22.23 & IE (Germanic) & Latin  \\
cop& 13.24  & Afro-Asiatic & Coptic \\
fo & 0.19 & IE (Germanic) & Latin\\
hu & 27.10 & Uralic & Latin \\
lt &  14.07 & IE (Baltic) & Latin\\
mt &  3.86 & Semitic & Latin \\
mr & 6.01 & IE (Indo-Aryan) & Devanagari\\
ta & 1.67 & Dravidian & Tamil \\
te & 0.15 & Dravidian & Telugu \\
wo &  2.99 & Niger-Congo & Latin \\
\end{tabular}
}
\caption{Information about the treebanks used.}
\end{center}
\end{table}

\section{Low-resource treebank sizes}
\label{appendix:lr-treebanks-size}
\begin{table}[!htpb]
\begin{center}
\small{
\begin{tabular}{l|c}
 & \# Sentences\\
\hline
Afrikaans\textsubscript{AfriBooms} & 1\,315  \\
Coptic\textsubscript{Scriptorium}     &   1\,089 \\
Faroese\textsubscript{FarPaHC}   &    1020\\
Hungarian\textsubscript{Szeged}  &   910 \\
Lithuanian\textsubscript{HSE}   &  153 \\
Maltese\textsubscript{MUDT}   &  1\,123    \\
Marathi\textsubscript{UFAL}  &   373\\
Tamil\textsubscript{TTB}  &   400 \\
Telugu\textsubscript{MTG}  &   1\,051 \\
Wolof\textsubscript{WTB}  &  1\,188 \\
\end{tabular}
}
\caption{Number of training sentences for the low-resource treebanks used.}
\end{center}
\end{table}

\end{document}